\documentclass[11pt]{article}

\usepackage[margin=1in]{geometry}
\usepackage[T1]{fontenc}
\usepackage{times}
\usepackage{amsmath,amssymb,amsthm,mathtools}
\usepackage{booktabs,multirow,array}
\usepackage{graphicx}
\usepackage{xcolor}
\usepackage{xspace}
\usepackage[numbers,sort&compress]{natbib}
\usepackage[colorlinks,citecolor=blue!60!black,linkcolor=blue!60!black,urlcolor=blue!60!black]{hyperref}
\usepackage[capitalise,noabbrev]{cleveref}

\newtheorem{proposition}{Proposition}

\theoremstyle{remark}

\allowdisplaybreaks[1]
\newcommand{\method}{\textsc{ELPSD}\xspace}

\newcommand{\nloops}{T}                              %
\newcommand{\params}{\theta}                         %
\newcommand{\baseparams}{\theta_0}                   %
\newcommand{\Ploop}[2]{P^{(#1)}_{#2}}                %
\newcommand{\student}{\pi_{\params}}                 %
\newcommand{\studentold}{\pi_{\params_{\mathrm{old}}}} %
\newcommand{\teacher}{\pi_{\mathrm{T}}}              %
\newcommand{\lbase}[1]{\ell_{#1}}                    %
\newcommand{\cdw}{w}                                 %
\newcommand{\refloop}{r}                             %
\newcommand{\loopdir}{f}                             %
\newcommand{\stuloop}{k}                             %

\newcommand{\prompt}{x}
\newcommand{\resp}{y}                                %
\newcommand{\prefix}{s}                              %
\newcommand{\groupsize}{G}                           %
\newcommand{\nresp}{N}                               %
\newcommand{\reward}{R}                              %
\newcommand{\adv}{\hat{A}}                           %
\newcommand{\vocab}{\mathcal{V}}                     %

\newcommand{\KL}[2]{\mathrm{KL}\!\left(#1\,\middle\|\,#2\right)}
\newcommand{\Renyi}[3]{D_{#1}\!\left(#2\,\middle\|\,#3\right)} %
\newcommand{\E}{\mathbb{E}}

\DeclareMathOperator{\softmax}{softmax}

\newcommand{\Lelpd}{\mathcal{L}_{\mathrm{ELPSD}}}
\newcommand{\Lloopopd}{\mathcal{L}_{\mathrm{LoopOPD}}}

\title{The Teacher Is Already Inside:\\Self-Distilling Looped Language Models Beyond Their Last Loop}
\author{Linghua Zhang}
\date{}

\begin{document}
\maketitle

\begin{abstract}
Looped language models apply one shared transformer stack several times per token, and every loop is readable through the shared language-model head. Post-training them is hard. Reward-only reinforcement learning needs a verifier and learns from one sparse signal per response; spreading the reward over loops collapses the entropy of every loop; and cross-loop self-distillation, as in LoopOPD, can only teach intermediate loops, because the last loop cannot learn from its own frozen copy: the loss is zero at initialisation. We observe that a looped model already contains a policy that decodes better than its last loop: the next-token distribution extrapolated from the first loop through the last. \method (Extrapolated-Loop Policy Self-Distillation) turns this distribution into a self-distillation target: a frozen copy of the model computes it, and the model's own last loop is distilled towards it on its own rollouts, using the exact reverse KL at every visited prefix. Teacher and student are the same model, so \method needs no reward, no external teacher and no privileged context, and it leaves rollouts and decoding unchanged. We show that the objective pulls the last loop towards the base model's last loop with weight $1+\cdw$ and away from the base model's first loop with weight $\cdw$, and that the teacher must stay frozen, since an idealised teacher recomputed from the current model extrapolates without bound. We evaluate \method on Ouro-1.4B-Thinking across five math benchmarks against the base model, GRPO, RLTT and OPSD.
Without any reward, \method raises the average avg@8 from 61.15 to 63.69 and is the best method on all five benchmarks, ahead of GRPO (62.32), which uses a verifier, and the self-distillation method OPSD (61.77), which needs reference solutions.

\end{abstract}

\section{Introduction}\label{sec:intro}

Looped language models reuse one set of transformer parameters across several passes per token~\citep{universaltransformer,ouro}.
Ouro~\citep{ouro} applies the same stack $\nloops=4$ times per token, and because all loops share one language-model head, every loop $t$ can be read out as a next-token distribution $\Ploop{t}{\params}$.
Rollouts and deployment use the last loop, $\student=\Ploop{\nloops}{\params}$.
Reinforcement learning with verifiable rewards (RLVR), usually GRPO~\citep{grpo,deepseekr1}, is the main tool for improving the reasoning of such models after pre-training.

Existing methods leave the deployed loop poorly served.
GRPO needs a verifier and learns from one scalar per response; a group whose responses are all correct or all wrong gives no gradient.
On Ouro-1.4B-Thinking trained on DeepMath problems~\citep{deepmath}, 100 GRPO steps raise avg@8 over five math benchmarks by only 1.17 points (61.15 to 62.32).
RLTT~\citep{rltt} spreads the reward over loops and trains every loop on tokens sampled from the last loop.
Running the authors' released code in the same setup, loop-1 token entropy falls to 23\% of its initial value and is still falling at step 100, and avg@8 ends below the base model on all five benchmarks, by 10.99 points on average (\cref{sec:exp-main}); LoopOPD~\citep{loopopd} also reports RLTT below the model before RL in its own setup.
LoopOPD removes the reward and lets the model teach itself: the frozen last loop of the base model is the teacher, and an intermediate loop, loop~2, is the student, trained with reverse KL on rollouts sampled from loop~2.
The loop that is deployed, the last loop, therefore improves only indirectly, through the parameters it shares with loop~2, and training requires sampling from an intermediate loop.
LoopOPD cannot simply make the last loop its student: a last loop taught by its own frozen copy starts at zero loss and zero gradient, so training never moves.

Our goal is reward-free self-distillation: a dense signal that comes from the looped model itself and trains the deployed last loop directly on its own rollouts, with no external teacher and no privileged context.
Our key observation is that a looped model already contains a policy better than its last loop.
Cross-loop contrastive decoding (LoopCD)~\citep{loopcd} extrapolates the log-probabilities from loop~1 through loop~$\nloops$, $(1+\cdw)\log\Ploop{\nloops}{}-\cdw\log\Ploop{1}{}$, and without any training raises the AIME24 pass@1 of Ouro-1.4B-Thinking from 50.83 to 59.17, as reported by \citet{loopcd}.
\citet{loopcd} attribute most of this gain to re-ranking tokens rather than to a change of temperature.
Internalising this policy into the weights pays off in three ways: the trained model decodes with standard last-loop sampling on any serving stack (LoopCD needs a custom model class in our vLLM implementation and runs without \texttt{torch.compile}); the student learns the teacher's behaviour on the states it visits itself; and the distilled model is a new starting point for test-time LoopCD or RL.

Training the last loop towards this policy raises three challenges.
\textbf{C1:} the deployed last loop has no teacher above it inside the model; with $\cdw=0$ the teacher is its own frozen copy and the objective is degenerate.
\textbf{C2:} an extrapolated teacher recomputed from the current model extrapolates its own extrapolation; the contrast between loops grows geometrically and there is no stable fixed point.
\textbf{C3:} the extrapolated teacher amplifies tokens that loop~1 strongly disprefers, which the student rarely uses; forcing the student to cover them pulls it towards states it does not visit.

We propose \method (Extrapolated-Loop Policy Self-Distillation), in which a looped model distills itself: its last loop learns, on its own rollouts, the LoopCD distribution of a frozen copy of the same model.
Unlike self-distillation that gives the teacher privileged context, such as a reference solution~\citep{opsd}, the teacher here sees exactly the student's input; its advantage comes only from reading the model's own loops differently.
With $\lbase{t}=\log\Ploop{t}{\baseparams}(\cdot\mid\prefix)$ and the loop direction $\loopdir=\lbase{\nloops}-\lbase{\refloop}$, the teacher is
$\log\teacher=\log\softmax\big((1+\cdw)\lbase{\nloops}-\cdw\lbase{\refloop}\big)=\lbase{\nloops}+\cdw\loopdir-\log Z(\prefix)$,
with $\cdw=0.5$ and $\refloop=1$, the same formula as test-time LoopCD.
Extrapolating beyond the last loop gives the deployed loop a teacher that differs from it at initialisation (C1), and computing the teacher from the frozen base model gives distillation a fixed point, the teacher itself (C2).
The objective is the full-vocabulary reverse KL $\KL{\student}{\teacher}$ at every prefix of rollouts sampled from the student, which is mode-seeking and on-policy (C3).
Up to a constant, this objective equals $(1+\cdw)\KL{\student}{\Ploop{\nloops}{\baseparams}}-\cdw\KL{\student}{\Ploop{1}{\baseparams}}$: \method pulls the last loop towards the base model's last loop and pushes it away from the base model's first loop (\cref{prop:decomp}).

Our contributions are as follows.
\begin{itemize}
  \item \textbf{Method.} \method, on-policy self-distillation for looped language models: the last loop learns the LoopCD distribution of the model's own frozen initial weights; it is reward-free, needs no external teacher or privileged context, and keeps standard last-loop rollouts and decoding (\cref{sec:method}).
  \item \textbf{Analysis.} A contrastive decomposition of the objective and of the initial gap, which equals a difference between a Rényi divergence and a KL divergence between the base model's last and first loops; a proof that the last loop cannot learn from its own frozen copy; and an argument that the extrapolated teacher must be frozen (\cref{sec:method-analysis}).
  \item \textbf{Experiments.} A comparison of \method with the base model, GRPO, RLTT and OPSD on five math benchmarks (\cref{sec:exp}).
  Without any reward, \method raises the average avg@8 from 61.15 to 63.69 and is the best method on all five benchmarks, ahead of GRPO and the self-distillation baseline OPSD.
\end{itemize}

\section{Related Work}\label{sec:related}

\paragraph{Looped language models and their post-training.}
Universal Transformers loop one shared block with adaptive halting~\citep{universaltransformer}.
Ouro~\citep{ouro} pretrains this design on 7.7T tokens; its exploratory RLVR with DAPO and GRPO gave no significant gain over the SFT checkpoint.
RLTT~\citep{rltt} replaces the last-loop log-probability in the policy gradient by a weighted sum over loops with one shared outcome advantage, applied to tokens sampled from the last loop.
LSRL~\citep{lsrl} adds external-grader scores of decoded intermediate states of the Huginn recurrent model to the outcome reward, but gradients flow only through the final-depth output.
LoopOPD~\citep{loopopd} removes the reward: an intermediate loop, on its own samples, minimises the full-vocabulary reverse KL to the last loop of the frozen initial model, and D-LoopOPD refreshes this teacher from the current parameters under a stop-gradient.
\method is also cross-loop self-distillation, but changes which loop learns and where the teacher sits.
Its student is the deployed last loop, trained on unchanged last-loop rollouts, and its teacher lies beyond the last loop rather than at it.
Its extrapolated teacher must also stay frozen (\cref{prop:frozen}), whereas D-LoopOPD can refresh its teacher because that teacher is the terminal loop itself, not an extrapolation of it.

\paragraph{On-policy distillation and self-distillation.}
On-policy distillation (OPD)~\citep{gkd} trains a student on its own rollouts towards a teacher's next-token distribution.
KDRL~\citep{kdrl} adds reverse-KL distillation from a fixed external teacher to GRPO as $J_{\mathrm{GRPO}}-\beta D^{k2}_{\mathrm{KL}}$.
ExOPD~\citep{exopd} casts OPD as dense KL-regularised RL with equal reward and KL weights; a reward scale $\lambda>1$ extrapolates the optimum beyond the teacher, giving a target of the form $\pi\propto\pi_{\mathrm{ref}}(\pi_{\mathrm{teacher}}/\pi_{\mathrm{ref}})^{\lambda}$.
D$^2$-OPD~\citep{d2opd} bounds the extrapolated target's distance from the teacher by a per-prefix KL constraint, and EOPD~\citep{eopd} adds a top-$k$ forward KL at high-entropy teacher positions.
All four need a separate teacher model.
Self-distillation removes the second model. OPSD~\citep{opsd} uses privileged information instead: the frozen initial model, given the problem together with its reference solution, teaches the same model without the solution, through a full-vocabulary divergence on the student's rollouts. It therefore needs a worked solution for every training problem.
Our teacher has the form of ExOPD's extrapolated optimum with $\pi_{\mathrm{ref}}=\Ploop{\refloop}{\baseparams}$, $\pi_{\mathrm{teacher}}=\Ploop{\nloops}{\baseparams}$ and $\lambda=1+\cdw$ (\cref{eq:teacher}).
It extrapolates between two loops of one frozen model instead of between two models, so both readouts come from a single forward pass, and it needs neither a second model nor reference solutions. \method is thus self-distillation whose privileged view is extra computation, the model's own loops read through extrapolation, rather than extra context.

\paragraph{Contrastive and extrapolated decoding.}
Contrastive Decoding~\citep{cd} corrects a large expert's logits with a small amateur's, and DoLa~\citep{dola} contrasts the final layer with an earlier layer chosen per token.
LoopCD~\citep{loopcd} contrasts loops of a looped model, decoding from $(1+\cdw)z_{\nloops}-\cdw z_{1}$ where $z_t$ are loop-$t$ logits, and attributes most of its gain to re-ranking tokens rather than to a temperature change; on Ouro-1.4B-Thinking it raises AIME24 pass@1 from 50.83 to 59.17 without training.
These methods change only decoding, so every generated token needs the contrast and the serving stack must support it.
\method trains the last loop towards the LoopCD distribution of the frozen base model, so the trained model decodes with standard last-loop sampling.

\begin{table}[t]
\centering
\small
\caption{Post-training methods for looped language models. ``Loop 2'' is LoopOPD's intermediate student loop $\stuloop=2$; ``last loop'' is loop $\nloops$, the loop used at deployment.}
\label{tab:compare}
\setlength{\tabcolsep}{4pt}
\begin{tabular}{@{}l c p{4.6cm} l l c@{}}
\toprule
Method & Reward / verifier & Teacher & Student & Rollouts from & Teacher fixed \\
\midrule
GRPO~\citep{grpo} & yes & none & last loop & last loop & n/a \\
RLTT~\citep{rltt} & yes & none (reward spread over loops by exit distribution) & all loops & last loop & n/a \\
LoopOPD~\citep{loopopd} & no & frozen base, last loop & loop 2 & loop 2 & yes \\
D-LoopOPD~\citep{loopopd} & no & current terminal loop, stop-gradient & loop 2 & loop 2 & no (refreshed) \\
OPSD~\citep{opsd} & no & frozen base, reference solution in context & last loop & last loop & yes \\
\method (ours) & no & frozen base, LoopCD (loop 1 $\to$ loop $\nloops$, $\cdw=0.5$) & last loop & last loop & yes \\
\bottomrule
\end{tabular}
\end{table}

\section{Method}\label{sec:method}

\method is on-policy self-distillation for looped language models. The last loop is trained, on its
own rollouts, towards a fixed teacher computed by the same model: the cross-loop contrastive
decoding (LoopCD) distribution of its frozen initial weights. The objective contains no reward.

\subsection{Preliminaries}\label{sec:prelim}

\paragraph{Looped language models.}
A looped language model applies one shared stack of transformer layers $\nloops$ times to every
token~\citep{universaltransformer,ouro}; Ouro~\citep{ouro} uses $\nloops=4$. With $h^{(t)}(\prefix)$
the normalised hidden state after loop $t$ at the last position of a prefix $\prefix$, the shared LM
head $W$ gives one next-token distribution per loop:
\begin{equation}\label{eq:loopdist}
  \Ploop{t}{\params}(\cdot\mid\prefix) = \softmax\!\big(W\,h^{(t)}(\prefix)\big),
  \qquad t = 1,\dots,\nloops .
\end{equation}
We disable early exit, so the policy, which we also call the student, is the last loop,
$\student \coloneqq \Ploop{\nloops}{\params}$, the distribution sampled at deployment. We write
$\baseparams$ for the initial parameters and
$\lbase{t} \coloneqq \log \Ploop{t}{\baseparams}(\cdot\mid\prefix)$ for the base model's loop-$t$
log-probabilities over the vocabulary $\vocab$.

\paragraph{On-policy self-distillation.}
The policy before the update, $\studentold$, samples responses $\resp_i$ to a prompt $\prompt$,
which define the visited prefixes $\prefix_{i,j} = (\prompt, \resp_{i,<j})$; a teacher scores every
visited prefix and the student is trained to match it there~\citep{gkd,kdrl,exopd}. In
self-distillation the teacher is the model itself with a privileged view; in OPSD~\citep{opsd} the
frozen initial model additionally sees the reference solution.
LoopOPD~\citep{loopopd} applies this inside a looped model with an intermediate loop $\stuloop=2$ as
the student, the frozen base last loop as the teacher, and rollouts from
$\Ploop{\stuloop}{\params_{\mathrm{old}}}$:
\begin{equation}\label{eq:loopopd}
  \Lloopopd(\params) = \frac{1}{\nresp}\sum_{i=1}^{\nresp}\frac{1}{|\resp_i|}\sum_{j=1}^{|\resp_i|}
  \KL{\Ploop{\stuloop}{\params}(\cdot\mid\prefix_{i,j})}{\Ploop{\nloops}{\baseparams}(\cdot\mid\prefix_{i,j})}.
\end{equation}
The deployed last loop changes only through the parameters it shares with loop $\stuloop$.

\subsection{A teacher beyond the last loop}\label{sec:method-teacher}

LoopCD~\citep{loopcd} applies contrastive decoding~\citep{cd,dola} inside one looped model, with the
last loop as the strong model and a reference loop $\refloop$ as the weak one; in log space it
extrapolates past the last loop along the direction from loop $\refloop$ to loop $\nloops$. Without
training it raises the AIME24 pass@1 of Ouro-1.4B-Thinking from 50.83 to 59.17, mostly by re-ranking
tokens rather than by a change of temperature~\citep{loopcd}. Our teacher is this rule applied to the
frozen base parameters, at temperature $1$ and without truncation:
\begin{equation}\label{eq:teacher}
  \log\teacher(\cdot\mid\prefix)
  = \log\softmax\!\big((1+\cdw)\,\lbase{\nloops} - \cdw\,\lbase{\refloop}\big)
  = \lbase{\nloops} + \cdw\,\loopdir - \log Z(\prefix),
\end{equation}
with $\cdw\ge0$, the base model's loop direction $\loopdir = \lbase{\nloops} - \lbase{\refloop}$, and
$Z(\prefix) = \sum_{v\in\vocab}\Ploop{\nloops}{\baseparams}(v\mid\prefix)^{1+\cdw}
\Ploop{\refloop}{\baseparams}(v\mid\prefix)^{-\cdw}$. Both readouts come from one forward pass of the
frozen model, and the teacher never changes during training.

\paragraph{Why the teacher must lie beyond the last loop.}
With $\cdw = 0$ the teacher is the base model's last loop, which is exactly the student at
initialisation: the loss below and its gradient are zero at every prefix, and training does not
move. The last loop cannot learn from its own frozen copy. LoopOPD avoids this by making an
intermediate loop the student, which leaves the deployed loop without a direct signal. \method keeps
the deployed loop as the student and moves the teacher beyond it; for $\cdw>0$ the two differ at
initialisation by the gap in \cref{prop:decomp}.

\subsection{Objective}\label{sec:method-objective}

At every step $\studentold$ samples $\groupsize$ responses per prompt with the standard last-loop
rollout engine, $\nresp$ in total. The loss is the exact reverse KL from the student to the teacher
at every visited prefix, averaged over the tokens of each response and then over responses, the
normalisation of LoopOPD:
\begin{equation}\label{eq:elpd}
  \Lelpd(\params) = \frac{1}{\nresp}\sum_{i=1}^{\nresp}\frac{1}{|\resp_i|}\sum_{j=1}^{|\resp_i|}
  \KL{\student(\cdot\mid\prefix_{i,j})}{\teacher(\cdot\mid\prefix_{i,j})} .
\end{equation}
Each prefix contributes only its own KL, computed in closed form over the vocabulary, so the
gradient is exact: prefixes are sampled, next tokens are not, and no importance ratio or clipping is
needed (\cref{app:proofs}). We take one optimiser update per step. The verifier is never called by
the objective; we only log the accuracy of the training rollouts.

\paragraph{Reverse rather than forward KL.}
The reverse KL is mode-seeking: it penalises mass where the teacher has little, but does not force
the student to cover every token the teacher supports. This suits an extrapolated teacher. Tokens
that loop $\refloop$ strongly disprefers have large $\loopdir$, so the teacher can raise them far
above their probability under the last loop, which rarely samples them. The forward KL
$\KL{\teacher}{\student}$ would require the student to cover all of them.

\subsection{Analysis}\label{sec:method-analysis}

All statements hold at a fixed prefix, which we drop; proofs are in \cref{app:proofs}.

\begin{proposition}[Contrastive decomposition and initial gap]\label{prop:decomp}
For every distribution $\pi$ over $\vocab$,
\begin{equation}\label{eq:decomp}
  \KL{\pi}{\teacher}
  = (1+\cdw)\,\KL{\pi}{\Ploop{\nloops}{\baseparams}}
  - \cdw\,\KL{\pi}{\Ploop{\refloop}{\baseparams}} + \log Z ,
\end{equation}
with $\log Z = \cdw\,\Renyi{1+\cdw}{\Ploop{\nloops}{\baseparams}}{\Ploop{\refloop}{\baseparams}} \ge 0$,
where $\Renyi{\alpha}{p}{q} = \frac{1}{\alpha-1}\log\sum_{v}p(v)^{\alpha}q(v)^{1-\alpha}$ is the
R\'enyi divergence of order $\alpha$. At initialisation, $\student = \Ploop{\nloops}{\baseparams}$
and
\[
  \KL{\Ploop{\nloops}{\baseparams}}{\teacher}
  = \cdw\,\Big[\Renyi{1+\cdw}{\Ploop{\nloops}{\baseparams}}{\Ploop{\refloop}{\baseparams}}
  - \KL{\Ploop{\nloops}{\baseparams}}{\Ploop{\refloop}{\baseparams}}\Big] \ge 0 ,
\]
which is non-decreasing in $\cdw$ and zero if and only if $\cdw=0$ or
$\Ploop{\nloops}{\baseparams} = \Ploop{\refloop}{\baseparams}$.
\end{proposition}

By \cref{eq:decomp}, minimising $\Lelpd$ pulls the last loop towards the base model's last loop with
weight $1+\cdw$ and pushes it away from the base model's loop $\refloop$ with weight $\cdw$; $\log Z$
is constant in $\params$. The pull term has the larger weight, and the sum is a KL divergence,
bounded below by zero. At $\params=\baseparams$ the pull term has
zero gradient, so the first updates move the last loop away from loop $\refloop$. The initial gap is
large where the two base loops disagree and zero where they agree.

\begin{proposition}[A self-refreshed extrapolated teacher diverges]\label{prop:frozen}
Idealise training as rounds. Let $\pi^{(0)} = \Ploop{\nloops}{\baseparams}$. In round $k$ the
teacher is recomputed from the current last loop,
$\log\pi_{\mathrm{T}}^{(k)} = \log\softmax\big((1+\cdw)\log\pi^{(k)} - \cdw\,\lbase{\refloop}\big)$,
with loop $\refloop$ held at its base value, and distillation reaches it exactly,
$\pi^{(k+1)} = \pi_{\mathrm{T}}^{(k)}$. Then
$\log\pi^{(k)} = \lbase{\refloop} + (1+\cdw)^{k}\loopdir - \log Z_k$. For $\cdw>0$, $\pi^{(k)}$
concentrates on $\arg\max_{v}\loopdir(v)$, and the only fixed point of the round map,
$\Ploop{\refloop}{\baseparams}$, is repelling. With the frozen teacher, the minimiser of
$\KL{\pi}{\teacher}$ is $\teacher$ itself, a fixed point of distillation.
\end{proposition}

The limit concentrates on tokens where loop $\nloops$ most exceeds loop $\refloop$, however unlikely
they are under loop $\nloops$, so it is a degenerate policy rather than a sharper teacher. A frozen
teacher removes this feedback. The refreshed teacher of D-LoopOPD~\citep{loopopd} is the current last
loop itself, not an extrapolation, so its recursion has no factor larger than one. Our teacher also
has the form of ExOPD's extrapolated optimum~\citep{exopd} with $\pi_{\mathrm{ref}}$ the base loop
$\refloop$, $\pi_{\mathrm{teacher}}$ the base loop $\nloops$ and $\lambda = 1+\cdw$. Summed along a
response, the per-prefix reverse KLs give the sequence-level KL to LoopCD decoding in expectation
(\cref{app:proofs}).

\subsection{Implementation}\label{sec:method-impl}

The teacher is a frozen float32 copy of the initial model on every training GPU. Its
full-vocabulary distributions are too large to store with the batch, so for each micro-batch the
teacher runs its loop-wise forward without gradients and passes its loop-1 and loop-4 hidden states
and LM head into the student's forward, which computes the KL of \cref{eq:elpd} in chunks of
positions under activation checkpointing. Rollouts use the same vLLM~\citep{vllm} engine and settings as GRPO. We
use $\cdw=0.5$ and $\refloop=1$, the LoopCD setting for Ouro-1.4B-Thinking on math~\citep{loopcd}.

\section{Experiments}\label{sec:exp}

\subsection{Setup}\label{sec:exp-setup}

\paragraph{Training.}
All trained systems start from Ouro-1.4B-Thinking~\citep{ouro} (24 layers, $\nloops = 4$) and update
all parameters on 14,684 problems filtered from DeepMath~\citep{deepmath}. Each step samples 64
prompts $\times$ $\groupsize = 8$ responses ($\nresp = 512$) at temperature 1.0 and top-p 1.0 and
performs one optimiser update; all runs take 100 steps with learning rate $10^{-6}$, responses of at
most 16,384 tokens, no KL term to a reference model, no entropy bonus, early exit disabled and seed
42. A binary outcome reward $\reward$ enters only the objectives of GRPO and RLTT; for the other
systems we compute it only to log training accuracy. Rollouts come from the last loop in every
system. \Cref{app:details} lists all settings.

\paragraph{Compared systems.}
\emph{Base} is Ouro-1.4B-Thinking without training. \emph{GRPO}~\citep{grpo} trains the last loop
with the clipped surrogate on group-normalised advantages. \emph{RLTT}~\citep{rltt} runs the authors'
released code in its original REINFORCE form, with exit-distribution weights over loops.
\emph{OPSD}~\citep{opsd} distills the last loop, on its own rollouts, towards the frozen initial model that additionally sees the reference solution of each problem, with the full-vocabulary generalised Jensen--Shannon divergence ($\beta = 0.5$, the paper's default) and no reward.
\emph{\method} minimises \cref{eq:elpd} with $\cdw = 0.5$, $\refloop = 1$.

\paragraph{Evaluation.}
We evaluate the step-100 checkpoints on AIME24, AIME25, AIME26, MATH-500~\citep{math,lightman} and AMC23, five benchmarks with 630 problems in total. Every model runs 4 loops
and decodes from its last loop without contrastive decoding. We draw 8 samples
per problem (temperature 0.6, top-p 0.95, at most 16,384 tokens, seed 42) and report avg@8, the mean
accuracy over all samples.

\subsection{Main results}\label{sec:exp-main}

\begin{table}[t]
\centering
\caption{avg@8 on five mathematics benchmarks (Ouro-1.4B-Thinking, step 100, 4 loops,
8 samples, temperature 0.6). Avg.\ is the unweighted mean. Best per column in bold.}
\label{tab:main}
\begin{tabular}{lcccccc}
\toprule
Method & AIME24 & AIME25 & AIME26 & MATH-500 & AMC23 & Avg. \\
\midrule
Base    & 45.42 & 40.83 & 42.08 & 91.78 & 85.63 & 61.15 \\
GRPO    & 45.00 & 40.83 & 45.00 & 92.65 & 88.13 & 62.32 \\
RLTT    & 32.50 & 27.92 & 29.17 & 85.58 & 75.63 & 50.16 \\
OPSD    & 44.58 & 40.00 & 44.58 & 92.20 & 87.50 & 61.77 \\
\method & \textbf{47.50} & \textbf{41.25} & \textbf{46.67} & \textbf{93.05} & \textbf{90.00} & \textbf{63.69} \\
\bottomrule
\end{tabular}
\end{table}

\Cref{tab:main} reports avg@8. Reward-only training gives a small gain: GRPO raises the
average avg@8 from 61.15 to 62.32 ($+1.17$); per benchmark,
avg@8 changes by $-0.42$ (AIME24) to $+2.92$ (AIME26). Spreading the reward over loops falls
below the base model: RLTT lowers avg@8 on all five benchmarks, by 6.20 (MATH-500) to 12.92
(AIME24) points, to an average of 50.16 ($-10.99$). During its training, loop-1 entropy falls from 1.684 to 0.382 nats, 23\% of its start, and all four loops end between 0.36 and 0.41 nats, consistent with training intermediate loops on tokens sampled from the last loop. Reward-free self-distillation with privileged information helps only slightly: OPSD raises the average to 61.77 ($+0.62$) without a verifier, but it needs a reference solution for every training problem.

\method gives the best avg@8 on all five benchmarks without any reward. It decodes exactly like
Base, and raises the average from 61.15 to 63.69 ($+2.54$), by $+0.42$ (AIME25) to $+4.59$
(AIME26) per benchmark. It exceeds GRPO by 1.37 points on average and on every benchmark ($+0.40$ to
$+2.50$), although GRPO calls the verifier on all 512 rollouts of every step and \method never does.
It also exceeds OPSD by 1.92 points on average and on every benchmark ($+0.85$ to $+2.92$) without any
reference solution: in this setting the model's own extra loops are a better teacher than privileged
context.

\section{Conclusion}\label{sec:conclusion}

A looped language model computes, but does not sample from, a distribution beyond its last loop:
the extrapolation from an early loop through the last one. \method lets the model distill itself:
its last loop learns, on its own rollouts, this LoopCD distribution of its own frozen initial
weights, with no reward, verifier, external teacher or privileged context, and with unchanged
rollouts and decoding. The objective pulls the last loop
towards the base loop $\nloops$ and away from the base loop $1$ (\cref{prop:decomp}), and the teacher
must stay frozen, since a self-refreshed one compounds the contrast (\cref{prop:frozen}).
Without any reward, \method raises the average avg@8 of Ouro-1.4B-Thinking from 61.15 to 63.69 and
is the best method on all five benchmarks, ahead of GRPO, RLTT and OPSD.

\paragraph{Future work.}
\method is complementary to reward-based training: an outcome reward can be added on top of the
self-distillation term, giving a reinforcement learning objective anchored at the extrapolated teacher.
Schedules for $\cdw$ and $\refloop$ could extrapolate further as training proceeds, and because the
self-distilled model is again a looped model, test-time LoopCD can be applied to it in turn. Extending
\method to larger looped models and to domains beyond mathematics is a natural next step.

\bibliographystyle{plainnat}
\bibliography{refs}

\clearpage
\appendix
\crefalias{section}{appendix}
\section{Proofs}\label{app:proofs}

\paragraph{Proof of \cref{prop:decomp}.}
By \cref{eq:teacher}, $\log\teacher = (1+\cdw)\lbase{\nloops} - \cdw\lbase{\refloop} - \log Z$.
Splitting $\log\pi = (1+\cdw)\log\pi - \cdw\log\pi$ in $\E_{\pi}[\log\pi - \log\teacher]$ gives
\cref{eq:decomp}. Next,
\[
Z = \sum_{v}\Ploop{\nloops}{\baseparams}(v)^{1+\cdw}\Ploop{\refloop}{\baseparams}(v)^{-\cdw}
= \E_{\Ploop{\nloops}{\baseparams}}\big[(\Ploop{\nloops}{\baseparams}/\Ploop{\refloop}{\baseparams})^{\cdw}\big]
= \exp\big(\cdw\,\Renyi{1+\cdw}{\Ploop{\nloops}{\baseparams}}{\Ploop{\refloop}{\baseparams}}\big)
\]
by the definition with $\alpha = 1+\cdw$, and R\'enyi divergences are non-negative. Setting
$\pi = \Ploop{\nloops}{\baseparams}$ in \cref{eq:decomp} removes the first term. The R\'enyi
divergence is non-decreasing in its order and equals the KL divergence at order $1$, so the bracket
is non-negative and non-decreasing in $\cdw$, and so is its product with $\cdw$. The gap is zero if
and only if $\teacher = \Ploop{\nloops}{\baseparams}$, i.e.\ $\cdw\loopdir(v) = \log Z$ for all $v$.
For $\cdw>0$ this makes $\loopdir$ constant; both loops are normalised softmax outputs with full
support, so $\loopdir = 0$. \hfill$\square$

\paragraph{Proof of \cref{prop:frozen}.}
For $k=0$, $\log\pi^{(0)} = \lbase{\nloops} = \lbase{\refloop} + \loopdir$. If the claim holds for
$k$, then up to a constant
$\log\pi^{(k+1)} = (1+\cdw)\big(\lbase{\refloop} + (1+\cdw)^{k}\loopdir\big) - \cdw\,\lbase{\refloop}
= \lbase{\refloop} + (1+\cdw)^{k+1}\loopdir$. Hence for $v^\star$ maximising $\loopdir$ and
$\loopdir(v)<\loopdir(v^\star)$, the ratio $\pi^{(k)}(v)/\pi^{(k)}(v^\star)$ equals
$\Ploop{\refloop}{\baseparams}(v)/\Ploop{\refloop}{\baseparams}(v^\star)$ times
$\exp\big((1+\cdw)^{k}(\loopdir(v)-\loopdir(v^\star))\big)\to0$. A fixed point has
$\cdw(\log\pi - \lbase{\refloop})$ constant, so $\pi = \Ploop{\refloop}{\baseparams}$, and a
perturbation $g$ of $\log\pi - \lbase{\refloop}$ is mapped to $(1+\cdw)g$ up to a constant, so the
fixed point is repelling. \hfill$\square$

The tokens in $\arg\max_v\loopdir(v)$ are those where loop $\nloops$ most exceeds loop $\refloop$,
however unlikely they are under loop $\nloops$, so the limit is a degenerate policy rather than a
sharper teacher.

\paragraph{Exact gradient of \cref{eq:elpd}.}
At a prefix $\prefix$, using $\sum_v \student(v)\nabla_\params\log\student(v)=0$,
\[
\nabla_{\params}\KL{\student}{\teacher} = \sum_{v\in\vocab}\student(v)\big[\log\student(v) -
\log\teacher(v)\big]\nabla_{\params}\log\student(v).
\]
Prefixes are sampled, next tokens are not, so no importance ratio or clipping is needed.

\paragraph{Sequence-level view.}
With $\student(\resp\mid\prompt)$ and $\teacher(\resp\mid\prompt)$ the products of per-token
distributions (the latter is LoopCD decoding of the base model at temperature 1), the expected sum
of per-prefix reverse KLs along a response from $\student$ equals
$\KL{\student(\resp\mid\prompt)}{\teacher(\resp\mid\prompt)}$. As in LoopOPD and KDRL~\citep{kdrl},
each prefix contributes its own KL term.

\paragraph{Relation to ExOPD.}
$\Ploop{\refloop}{\baseparams}(\Ploop{\nloops}{\baseparams}/\Ploop{\refloop}{\baseparams})^{1+\cdw}
= (\Ploop{\nloops}{\baseparams})^{1+\cdw}(\Ploop{\refloop}{\baseparams})^{-\cdw}$, so the teacher has
the form of ExOPD's extrapolated optimum~\citep{exopd} at every prefix, with
$\pi_{\mathrm{ref}} = \Ploop{\refloop}{\baseparams}$, $\pi_{\mathrm{teacher}} =
\Ploop{\nloops}{\baseparams}$ and $\lambda = 1+\cdw$.

\section{Experimental details}\label{app:details}

\begin{table}[h]
\centering
\caption{Settings shared by all trained systems.}
\label{tab:hp-shared}
\small
\begin{tabular}{p{0.36\textwidth}p{0.58\textwidth}}
\toprule
Setting & Value \\
\midrule
Initial model & Ouro-1.4B-Thinking, 24 layers, $\nloops = 4$ loops, full-parameter training \\
Training data & 14,684 filtered DeepMath problems, binary outcome reward \\
Prompts $\times$ responses per step & $64 \times 8 = 512$ \\
Optimiser updates per step & 1 (mini-batch = 64 prompts, 1 epoch) \\
Training steps & 100 \\
Learning rate & $10^{-6}$, 10 warm-up steps, then constant \\
Optimiser & AdamW~\citep{adamw}, betas $(0.9, 0.99)$, weight decay $0.1$ \\
Gradient clipping & global norm $0.1$ \\
KL term to a reference model & none \\
Entropy bonus & none \\
Max prompt / response length & 2,048 / 16,384 tokens \\
Rollout sampling & temperature 1.0, top-p 1.0, vLLM in bf16 \\
Early exit & disabled (log-probabilities always from loop 4) \\
Seed (training and data order) & 42 \\
Hardware & one node, 8 GPUs \\
\bottomrule
\end{tabular}
\end{table}

\paragraph{Method-specific settings.}
GRPO: clipped surrogate (clip 0.2, dual clip 3) on group-normalised advantages, last loop.
RLTT: authors' code, REINFORCE on $\adv_i\sum_t q_t\log\Ploop{t}{\params}$ with the exit distribution
$q_t$, exit gate trained. OPSD: frozen initial model with the reference solution in its context as teacher, last loop as student on its own rollouts, full-vocabulary generalised JSD with $\beta = 0.5$ (the defaults of \citet{opsd}). \method: $\cdw = 0.5$, $\refloop = 1$, exact reverse KL,
per-response token mean; the frozen float32 teacher is held unsharded on every GPU (about
5.6\,GB), runs its loop-wise forward per micro-batch without gradients, and passes its loop-1 and
loop-4 hidden states and LM head into the student's forward, where the KL is computed in chunks of
positions under activation checkpointing.

\end{document}